\RequirePackage[l2tabu, orthodox]{nag}
\documentclass[11pt]{article}

\usepackage[T1]{fontenc}

\usepackage{soul} 

\usepackage{garamond}  
\renewcommand{\rmdefault}{ugm}

\usepackage[bitstream-charter]{mathdesign}
\usepackage{amsmath}
\usepackage[scaled=0.92]{PTSans}

\usepackage[
  paper  = letterpaper,
  left   = 1.65in,
  right  = 1.65in,
  top    = 1.0in,
  bottom = 1.0in,
  ]{geometry}

\usepackage[dvipsnames]{xcolor}
\definecolor{shadecolor}{gray}{0.9}

\usepackage[final,expansion=alltext]{microtype}
\usepackage[english]{babel}
\usepackage[parfill]{parskip}
\usepackage{afterpage}
\usepackage{framed}

{\endMakeFramed}

\usepackage{lineno}

\usepackage{ragged2e}

\newcounter{parcount}

\usepackage{graphicx}
\usepackage[labelfont=bf]{caption}

\usepackage{booktabs}
\usepackage{multirow}

\usepackage[algoruled]{algorithm2e}
\usepackage{listings}
\usepackage{fancyvrb}
\fvset{fontsize=\normalsize}

\usepackage{natbib}

\usepackage[acronym,nowarn]{glossaries}
\makeglossaries

\definecolor{tangerine}{rgb}{0.95, 0.52, 0.0}
\definecolor{palebrown}{rgb}{0.6, 0.46, 0.33}
\definecolor{peru}{rgb}{0.8, 0.52, 0.25}
\usepackage[colorlinks,linktoc=all]{hyperref}
\usepackage[all]{hypcap}
\hypersetup{citecolor=peru}
\hypersetup{linkcolor=peru}
\hypersetup{urlcolor=peru}

\usepackage[nameinlink]{cleveref}
\creflabelformat{equation}{#1#2#3}
\crefname{equation}{eq.}{eqs.}  
\Crefname{equation}{Eq.}{Eqs.}

\lstdefinestyle{mystyle}{
    commentstyle=\color{OliveGreen},
    keywordstyle=\color{BurntOrange},
    numberstyle=\tiny\color{black!60},
    stringstyle=\color{MidnightBlue},
    basicstyle=\ttfamily,
    breakatwhitespace=false,
    breaklines=true,
    captionpos=b,
    keepspaces=true,
    numbers=left,
    numbersep=5pt,
    showspaces=false,
    showstringspaces=false,
    showtabs=false,
    tabsize=2
}
\usepackage[colorinlistoftodos,
           prependcaption,
           textsize=small,
           backgroundcolor=yellow,
           linecolor=lightgray,
           bordercolor=lightgray]{todonotes}

\usepackage{amsthm}  

\usepackage{bm}

\usepackage[colorinlistoftodos,
           prependcaption,
           textsize=small,
           backgroundcolor=yellow,
           linecolor=lightgray,
           bordercolor=lightgray]{todonotes}

\usepackage{graphicx}
\usepackage{caption}
\usepackage{subcaption}
\usepackage{wrapfig}

\usepackage{booktabs}
\usepackage{arydshln} 
\usepackage{multirow}
\usepackage{nicematrix}

\usepackage{listings}
\usepackage{fancyvrb}
\fvset{fontsize=\normalsize}

\usepackage[nameinlink]{cleveref}
\creflabelformat{equation}{#1#2#3}
\crefname{equation}{eq.}{eqs.}  
\Crefname{equation}{Eq.}{Eqs.}

\usepackage{listings}
\lstdefinestyle{alp_style}{
    commentstyle=\color{OliveGreen},
    numberstyle=\tiny\color{black!60},
    stringstyle=\color{BrickRed},
    basicstyle=\ttfamily\scriptsize,
    breakatwhitespace=false,
    breaklines=true,
    captionpos=b,
    keepspaces=true,
    numbers=none,
    numbersep=5pt,
    showspaces=false,
    showstringspaces=false,
    showtabs=false,
    tabsize=2
}

\usepackage{capt-of}

\usepackage{lipsum}

\theoremstyle{remark}
\newtheorem*{lemma*}{Lemma}

\usepackage{authblk}
\usepackage{enumitem} 
\usepackage{algorithmic}
\usepackage{tabularx}
\usepackage{booktabs}
\usepackage{pifont}
\usepackage[T1]{fontenc}
\usepackage{multirow}
\usepackage{threeparttable}
\usepackage{tcolorbox}

\title{\textbf{Vendi-RAG: Adaptively Trading-Off Diversity And Quality Significantly Improves Retrieval Augmented Generation With LLMs}}

\author[1, 3]{Mohammad R. Rezaei}
\author[2, 3]{Adji Bousso Dieng}
\affil[1]{Institute of Biomedical Engineering, University of Toronto}
\affil[2]{Department of Computer Science, Princeton University}
\affil[3]{\href{https://vertaix.princeton.edu/}{Vertaix}}

\makeglossary 

\begin{document}
\maketitle

\begin{abstract}
\noindent Retrieval-augmented generation (RAG) enhances large language models (LLMs) for domain-specific question-answering (QA) tasks by leveraging external knowledge sources. However, traditional RAG systems primarily focus on relevance-based retrieval and often struggle with redundancy, especially when reasoning requires connecting information from multiple sources. This paper introduces \textbf{Vendi-RAG}, a framework based on an iterative process that jointly optimizes retrieval diversity and answer quality. This joint optimization leads to significantly higher accuracy for multi-hop QA tasks. Vendi-RAG leverages the Vendi Score (VS), a flexible similarity-based diversity metric, to promote semantic diversity in document retrieval.~It then uses an LLM judge that evaluates candidate answers, generated after a reasoning step, and outputs a score that the retriever uses to balance relevance and diversity among the retrieved documents during each iteration. Experiments on three challenging datasets---HotpotQA, MuSiQue, and 2WikiMultiHopQA---demonstrate Vendi-RAG's effectiveness in multi-hop reasoning tasks. The framework achieves significant accuracy improvements over traditional single-step or multi-step RAG approaches, with accuracy increases reaching +4.2\% on HotpotQA, +4.1\% on 2WikiMultiHopQA, and +1.3\% on MuSiQue compared to Adaptive-RAG, the current best baseline. The benefits of Vendi-RAG are even more pronounced as the number of retrieved documents increases. Finally, we evaluated Vendi-RAG across different LLM backbones, including GPT-3.5, GPT-4, and GPT-4o-mini, and observed consistent improvements, demonstrating that the framework's advantages are model-agnostic.\\

\end{abstract}

\section{Introduction}

\begin{figure*}[!t]
\centering
\includegraphics[width=.9\linewidth]{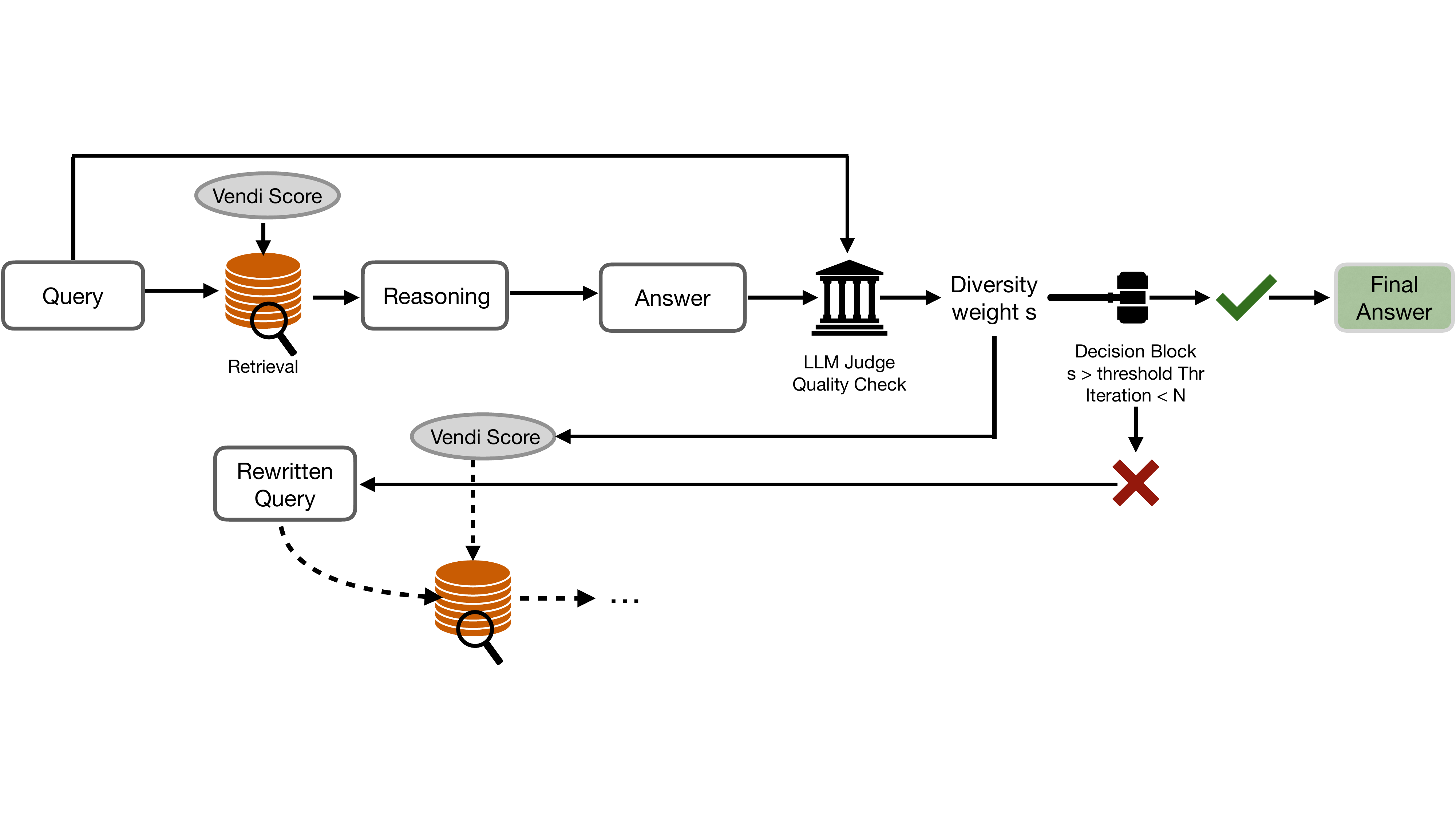}
\caption{The process begins with an initial retrieval step, where a diverse set of documents is retrieved using the Vendi Score, ensuring broad semantic coverage. Next, leveraging a reasoning step to construct a coherent path to the final answer, the LLM generates an answer, which then undergoes quality assessment by an LLM judge. Based on the answer quality, the retriever is adjusted to balance diversity and relevance: high-quality answers limit the emphasis on diversity, while low-quality answers prompt the retriever to prioritize diversity more heavily. This adjustment is controlled by an adaptive parameter, $s$, which is updated over iterations. The process continues until the answer quality reaches an optimal threshold, denoted by Thr. Finally, the highest-quality responses and documents are selected, ensuring both diversity and accuracy.}
\label{fig:vendi-rag}
\end{figure*}

Retrieval-augmented generation (RAG) has emerged as a transformative framework for enhancing the performance of large language models (LLMs) in domain-specific tasks such as question-answering (QA). By retrieving relevant information from external sources beyond the training set, RAG enables LLMs to answer specialized queries more effectively \cite{achiam2023gpt, team2023gemini, jiang2024mixtral}. This approach has been particularly successful in single-hop QA, where a question can be answered using information from a single document \cite{raiaan2024review, kwiatkowski2019natural}. For instance, answering a question such as "Who wrote the novel Frankenstein?" only requires retrieving relevant information from a single document containing this fact.

However, multi-hop QA introduces significantly greater complexity. Finding the correct answer to queries in multi-hop QA requires reasoning across multiple sources \citep{press2022measuring, tang2024multihop}. For instance, answering "Which city is the capital of the African country where Mount Kilimanjaro is located?" necessitates first identifying that Mount Kilimanjaro is in Tanzania, and then determining that Dodoma is the capital of Tanzania. This process involves not only retrieving information from multiple documents but also synthesizing these different sources effectively to form an accurate answer, which greatly increases the complexity of both retrieval and reasoning and leads to redundancy.

To address these challenges, iterative RAG pipelines have been developed. These pipelines refine the retrieval process through repeated modifications and re-querying of retrieved documents, aiming to resolve ambiguities and improve relevance. Notable examples include Adaptive-RAG~\citep{lewis2020retrieval}, which controls the number of iterations of the pipeline including the retrieval process and modifying the queries based on a classification model's assessment of the input query, Self-RAG\cite{asai2023self}, which incorporates iterative self-reasoning, and IROC \cite{trivedi2022interleaving}, which progressively refines retrieval to optimize the final answer \cite{wei2022chain, wang2024chainofthought}.

Despite their success, iterative RAG methods typically rely solely on relevance-based retrieval, which focuses on the similarity between the query and dataset entries. This approach presents a fundamental limitation: it does not actively manage the diversity and quality of the retrieved information to properly address the query. More complex queries require diverse retrieval. We therefore propose a novel retrieval method called \emph{Vendi retrieval} to address the limitation of existing retrieval pipelines. Vendi retrieval leverages the Vendi Score (VS) to enhance the diversity of retrieved documents while accounting for retrieval quality through a simple weighting mechanism. 

Building on Vendi retrieval, we propose an iterative RAG pipeline called Vendi-RAG that balances diversity and quality. More specifically, the pipeline is as follows: an initial set of candidate documents is retrieved. Based on these retrieved documents, the system generates chain-of-thought (CoT) reasoning steps. Using these reasoning steps and retrieved documents, the LLM then generates candidate answers. An LLM-based evaluator then assesses these candidates for relevance, coherence, and completeness. The highest-scoring answer is selected as the final response. If the answer does not meet the quality threshold, the Vendi retrieval process dynamically adjusts the balance between diversity and relevance in document selection, ensuring broader semantic exploration or increased specificity as needed. This iterative refinement continues until a high-quality response is achieved. Figure \ref{fig:vendi-rag} provides a detailed overview of the Vendi-RAG framework. 

We evaluated the Vendi retrieval process and Vendi-RAG on three challenging multi-hop QA datasets, HotpotQA \citep{yang2018enhancing}, MuSiQue \citep{trivedi2022interleaving}, and 2WikiMultiHopQA \citep{ho2020constructing}. To assess the Vendi retrieval method we measured the diversity of retrieved documents on the three datasets using two different diversity metrics, the VS and the max pairwise distance (MPD). We found that the Vendi retrieval process yields more diverse documents compared to the baselines according to both metrics. Second, we evaluated Vendi-RAG in terms of several performance metrics, looking at both accuracy and diversity. The results showed that Vendi-RAG substantially improves response accuracy, outperforming existing RAG approaches. Using GPT-3.5 as the LLM backbone, Vendi-RAG demonstrated significant accuracy gains across all datasets, with accuracy increases reaching +4.2\% on HotpotQA, +4.1\% on 2WikiMultiHopQA, and +1.3\% on MuSiQue compared to Adaptive-RAG, the best baseline. Notably, the accuracy improvement remained consistent across different LLM backbones---GPT-4o, GPT-4o-mini, and GPT-3.5---indicating that Vendi-RAG's advantages are model-agnostic. Additionally, our experiments with varying numbers of retrieved documents---beyond the standard two-document setting---showed that Vendi-RAG maintained its superior performance, especially as the number of retrieved documents increased. This underscores the critical role of the Vendi retrieval process in handling complex retrieval scenarios. For instance, when retrieving ten documents from HotpotQA, Vendi-RAG outperformed Adaptive-RAG by 7.8\% in accuracy using GPT-4o-mini as the backbone LLM.

This work introduces a diversity-guided retrieval approach that optimizes both diversity and quality to address the challenges of multi-step reasoning in multi-hop QA. Our experimental results highlight the effectiveness of Vendi-RAG in enhancing retrieval diversity and response accuracy, underscoring its potential as a robust solution for complex multi-hop QA tasks.

\section{Related Work}
\label{sec:related}

There are three main approaches to QA: non-retrieval-based methods~\citep{petroni2019language}, single-step RAG~\citep{lewis2020retrieval}, and multi-step RAG~\citep{asai2023self}. Non-retrieval-based QA methods pass queries directly to an LLM and use its generated output as the answer, without consulting external sources. While efficient, these methods struggle with queries requiring external or up-to-date information and suffer from hallucinations on out-of-distribution queries~\citep{shuster2021retrieval}. Single-step RAG methods integrate external knowledge retrieved from a knowledge base (e.g., Wikipedia). These methods improve factual accuracy but are limited by retrieval noise and perform poorly in complex reasoning tasks~\citep{trivedi2022interleaving}. Multi-step RAG methods are designed for complex multi-hop queries~\citep{jeong2024adaptive,asai2023self,tang2024multihop}. They iteratively retrieve documents and refine answers until they converge on a final response. This iterative refinement approach enables reasoning across multiple sources but introduces computational overhead and is prone to error accumulation~\citep{jeong2024adaptive}.

\paragraph{Advances in multi-hop QA.} Recent improvements in multi-hop QA focus on question decomposition~\citep{radhakrishnan2023question}, chain-of-thought reasoning~\citep{wei2022chain,liu2024much}, and iterative retrieval~\citep{jeong2024adaptive,shao2023enhancing,yu2024auto}. Methods like ReCite~\citep{sun2022recitation} and IRCoT~\citep{trivedi2022interleaving} refine retrieval with progressive reasoning, while Self-RAG~\citep{asai2023self} adapts retrieval strategies based on query complexity. Decomposed prompting~\citep{khot2022decomposed} further enhances retrieval for complex queries~\citep{zhang2024accelerating}. MultiHop-RAG~\citep{tang2024multihop} integrates decomposition and retrieval pipelines but remains constrained by relevance-based retrieval, leading to redundancy and limited synthesis of diverse information. 

\paragraph{Vendi scoring.} The Vendi Score (VS)~\citep{friedman2023vendi} is a similarity-based diversity metric applied in machine learning~\citep{berns2023towards, pasarkar2023cousins, mousavi4924208vsi, nguyen2024quality, kannen2024beyond, jalali2024conditional, askari2024improving, rezaei2025alpha, bhardwaj2025robust}, chemistry~\citep{pasarkar2023vendi}, materials science~\citep{liu2024diversity}, and biology~\citep{pasarkar2025vendiscope}. Vendi-RAG integrates VS into retrieval, balancing diversity and quality beyond conventional ranking systems~\citep{carbonell1998use, slivkins2010learning}. Unlike standard relevance-based retrieval~\citep{guu2020retrieval}, this approach enhances robustness and accuracy in multi-hop QA by incorporating semantic diversity into document retrieval.

\section{Method}
\label{sec:method}

We now describe Vendi-RAG, including the novel retrieval process it uses.

\subsection{Vendi Retrieval}
Diversity in retrieved documents is essential for multi-hop QA, as it ensures broad semantic coverage, reduces redundancy, and incorporates multiple perspectives~\citep{sun2022recitation, carbonell1998use, thakur2021beir}. The most commonly used techniques for information retrieval are similarity search (SS)\citep{thakur2021beir} and maximal marginal relevance (MMR)\citep{carbonell1998use}. While SS retrieves documents based on their relevance to the query, it often results in redundant documents with high similarity. MMR attempts to balance relevance and novelty using pairwise comparisons, but it still struggles to capture global semantic diversity.

To address these limitations, we adopt a retrieval approach based on the Vendi Score (VS)~\citep{friedman2023vendi}, which explicitly quantifies semantic diversity in a set of documents. Let $\mathcal{D} = {d_1, \dots, d_n}$ be a set of retrieved documents, and let $k(\cdot, \cdot)$ be a positive semi-definite similarity kernel such that $k(d_i, d_i) = 1$ for all $i$. Let $K$ be the similarity matrix with entries $K_{ij} = k(d_i, d_j)$. The Vendi Score is
\begin{align}
\text{VS}_k(\mathcal{D}) = \exp\left(-\sum_{i=1}^n \lambda_i \log \lambda_i\right),
\end{align}
where $\lambda_1, \dots, \lambda_n$ are the eigenvalues of the normalized kernel matrix $K / n$. As shown by \citet{friedman2023vendi}, $\text{VS}_k(\mathcal{D})$ reflects the effective number of unique documents in $\mathcal{D}$, attaining its maximum value $n$ when all documents are orthogonal (fully diverse) and its minimum value 1 when all documents are identical.

While optimizing for diversity is important—especially for complex, multi-faceted queries—it must be balanced with query relevance. To achieve this, we define the Vendi Retrieval Score (VRS) as a convex combination of semantic diversity and similarity-based relevance:
\begin{align}
\text{VRS} = s \cdot \text{VS}_k(\mathcal{D}) + (1-s) \cdot \text{SS}(q, \mathcal{D}),
\end{align}
where $s \in [0, 1]$ is a tunable parameter that controls the trade-off between diversity and relevance. The similarity score $\text{SS}(q, \mathcal{D})$ is computed using dense vector representations of the query $q$ and the documents in $\mathcal{D}$, typically obtained from transformer-based encoders. This ensures semantic matching beyond surface-level lexical overlap.

It is important to note that while the Vendi Score $\text{VS}_k(\mathcal{D})$ is computed solely based on the retrieved documents and their pairwise similarities, query relevance is introduced in the initial retrieval step: the candidate set $\mathcal{D}$ is selected using similarity search with respect to the query $q$. Thus, the formulation in Equation (2) balances document-level diversity and query-level relevance, where a higher value of $s$ favors diverse content, and a lower value prioritizes semantic similarity to the query. In this way, the VSR addresses the dual objectives of reducing redundancy and maintaining relevance, providing a principled and flexible framework for multi-hop document selection.
\subsection{Vendi-RAG}  
We integrate the Vendi retrieval process into a flexible RAG pipeline that balances diversity and relevance for improved performance on multi-hop QA. 

\paragraph{1. Initial retrieval.} The process begins by retrieving a set of documents using Vendi retrieval. This first step prioritizes broad semantic coverage (we set $s = 0.8$ initially in all our experiments), ensuring that the retrieved documents capture multiple perspectives and to prevent recovering semantically redundant documents. This initial diversity is particularly critical for multi-hop QA, where synthesizing information from varied sources is essential to accurately answering the query.
    
\paragraph{2. Reasoning generation.} Based on the retrieved documents, the system generates CoT reasoning steps. These intermediate reasoning steps help contextualize the retrieved information, building a coherent pathway to the final answer.

\paragraph{3. Candidate answer generation.} Using the reasoning steps and retrieved documents, the LLM generates candidate answers. These proposed answers are evaluated to determine their quality and completeness.

\paragraph{4.~Quality evaluation.} An LLM judge assesses the candidate answers. This evaluation considers factors such as coherence, relevance, and alignment with the query. A quality score $Q_t$ is produced at the end of this quality-check. Here $t$ is used to indicate the iteration step. 

\paragraph{5. Dynamic adjustment of the VRS.} Based on the quality score $Q_t$, the parameter $s$ is adjusted dynamically. We denote by $s_t$ the value of the parameter $s$ at the $t^{\text{th}}$ iteration. It controls the trade-off between diversity (via VS) and relevance (via similarity search). If  $Q_t$ is low, $s_t$ should be increased, to prioritize greater diversity in the subsequent retrievals. This ensures broader semantic exploration, which is beneficial for refining answers in cases where the retrieved information is already relevant but lacks coverage. Conversely, if $Q_t$ is high, $s_t$ should be decreased to focus more on relevance, retrieving documents that are closely aligned with the query to address potential gaps in specificity. We therefore define $s_t$ as
\begin{align}
    s_t &= f(Q_{t-1})=1- \frac{Q_{t-1}}{\max(Q_{t-1})},
\end{align}
where $f$ is a simple linear function that maps \( Q_{t-1} \) to the interval \([0,1]\), ensuring that higher quality scores correspond to lower diversity scores.
\paragraph{6. Iterative refinement.} The retrieval and reasoning steps are repeated iteratively, with adjustments to $s$ dynamically balancing diversity and relevance at each stage. This process continues until the desired answer quality is reached, ensuring that the system converges on an optimal set of documents and reasoning steps.
\paragraph{7. Final answer selection.} Once the iterative refinement process is complete, the final set of documents and answers are selected based on their quality scores. This ensures that the output reflects both broad semantic coverage and high-quality, relevant information. Algorithm~\ref{alg:vendi-rag} summarizes the procedure.
\paragraph{Why Adjusting $s$ Matters:}
The dynamic adjustment of $s$ is essential for balancing diversity and relevance during retrieval. High diversity enables exploration of different facets of complex queries, especially in multi-hop QA, where synthesizing information from multiple sources is crucial. However, too much diversity can introduce noise, while excessive focus on relevance risks redundancy and limits comprehensive reasoning.

Vendi-RAG addresses this by adapting $s$ based on retrieval quality: when the quality score is high, it reduces $s$ to promote exploration of additional, semantically diverse documents; when quality is low, it increases $s$ to prioritize more directly relevant documents. This adaptive retrieval strategy allows Vendi-RAG to dynamically adjust to the needs of each query and reasoning step, improving both the breadth and precision of generated answers. Unlike traditional RAG systems with fixed retrieval policies, Vendi-RAG's flexibility ensures richer, more contextually appropriate responses.
\paragraph{Performance characteristics.}
In practice, Vendi-RAG exhibits distinctive performance patterns that reflect its sophisticated design. The system naturally adapts its computational effort to query complexity, requiring more iterations for intricate multi-hop queries while converging quickly for simpler ones. Though the computational overhead exceeds that of basic RAG systems, the improved retrieval quality often results in better final answers. The system maintains reasonable scalability with document corpus size, as the primary computational bottleneck—eigenvalue computation—depends on the number of retrieved documents rather than the total corpus size. These characteristics make Vendi-RAG particularly suitable for complex tasks such as multi-hop QA. 
\begin{algorithm}[t]
\DontPrintSemicolon
\caption{Vendi-RAG Inference Pipeline}
\label{alg:vendi-rag}
Inputs: Query $q$, knowledge base $\mathcal{D}$, \# iterations $N$, quality threshold $\tau$\;

Initialize query $q_1 \gets q$, initialize parameter $s_1 \gets 0.8$\;

\For{$i = 1, \dots, N$}{

    Compute scores: $\text{VSR}_i = s_i \cdot \text{VS}_k(\mathcal{D}) + (1 - s_i) \cdot \text{SS}(q, \mathcal{D})$\;

    Get the document set: $D_i \gets \text{Vendi-Retriever}(\text{VSR}_i; \mathcal{D})$\;
    
    Generate reasoning: $r_i \gets \text{CoT}(q, D_i)$\;
    
    Generate answers: $\hat{a}_i \gets \text{LLM}(q, D_i, r_{1:i})$\;
    
    Evaluate quality: $Q_i \gets \text{LLM-Judge}(\hat{a}_i)$\;

    If $Q_i \geq \tau$ then return $\hat{a}_i$\;

    Else update $q$ and s: $q_{i+1} \gets \text{RewriteQuery}(q_i, \hat{a}_i, r_i)$ and $s_{i+1} \gets f(Q_{i})$\;
}
\Return $\hat{a}_N$ \tcp*[f]{Return the best answer after $N$ iterations.}
\end{algorithm}

\section{Experiments} \label{sec:experiments}
In this section, we present a comprehensive evaluation of Vendi-RAG on multi-hop QA tasks. We begin by analyzing the effectiveness of the Vendi retrieval strategy in enhancing retrieval diversity. We then evaluate the full Vendi-RAG pipeline, highlighting its ability to handle complex queries requiring multi-step reasoning, and compare its performance against several strong baselines. All experiments are conducted on three challenging multi-hop QA benchmark datasets: {MuSiQue} \citep{trivedi2022interleaving}, {HotpotQA} \citep{yang2018enhancing}, and {2WikiMultiHopQA} \citep{ho2020constructing} (see Appendix~\ref{app:datasets} for additional dataset details).

\begin{table}[t]
\small
\centering
\begin{tabular}{llcc}
\toprule
Dataset & Method & VS & MPD \\
\midrule
\multirow{3}{*}{MuSiQue} 
& Adaptive Retrieval & 6.13 & 1.25 \\
& Adaptive-RAG        & 6.55 & 1.42 \\
& Vendi-RAG           & \textbf{7.12} & \textbf{1.95} \\
\midrule
\multirow{3}{*}{HotpotQA}
& Adaptive Retrieval & 4.95 & 1.10 \\
& Adaptive-RAG        & 5.21 & 1.31 \\
& Vendi-RAG           & \textbf{6.82} & \textbf{1.89} \\
\midrule
\multirow{3}{*}{2WikiMHQA}
& Adaptive Retrieval & 5.34 & 1.32 \\
& Adaptive-RAG        & 5.81 & 1.45 \\
& Vendi-RAG           & \textbf{6.68} & \textbf{1.78} \\
\bottomrule
\end{tabular}
\caption{Retrieval diversity (Vendi Score (VS) and Max Pairwise Distance (MPD)) across datasets and methods. Vendi-RAG achieves higher diversity than baselines.}
\label{tab:vs_mpd_comparison}
\end{table}
\paragraph{Sensitivity Analysis of the VSR Process.}
To evaluate the robustness of the VSR process and understand its impact on retrieval diversity, we conducted a sensitivity analysis focusing on how varying the parameter \( s \) affects document ranking order within a vector database. This analysis helps elucidate the trade-off between retrieval precision and diversity, which is crucial for enhancing multi-hop reasoning performance. The sensitivity analysis was performed using 100 randomly sampled queries from the dataset to ensure a comprehensive evaluation covering a diverse range of query types and complexity levels. Our primary objective was to investigate how different values of \( s \) influence retrieval diversity and ranking consistency.

We evaluated the retrieval pipeline across multiple \( s \) values ranging from 0.0 to 1.0, incremented in small steps to capture granular variations in retrieval performance. Setting \( s = 0.0 \) serves as a baseline representing a pure similarity search scenario, where retrieval relies exclusively on cosine similarity or dot product between embeddings, without any emphasis on diversity. This baseline provides a reference point for measuring the impact of increasing \( s \) on retrieval diversity.
To quantify deviations from the baseline, we employed two complementary ranking comparison metrics:
\begin{itemize}
    \item \textbf{Kendall’s \( \tau \):} Measures the rank order similarity between two lists by evaluating concordant and discordant pairs. Higher \( \tau \) values indicate stronger similarity to the baseline, while lower values reflect greater diversity introduced by increasing \( s \). 
    \item \textbf{Spearman’s Rank Correlation \( \rho \):} Assesses the monotonic relationship between two rankings by considering both orderings and positional shifts. Lower \( \rho \) values signal substantial deviation from the baseline, indicating increased diversity through higher \( s \) values.
\end{itemize}
The results of the sensitivity analysis are presented in Table~\ref{tab:sensitivity_analysis}. As \( s \) increases from 0.0 to 1.0, both Kendall’s \( \tau \) and Spearman’s \( \rho \) decrease progressively, demonstrating that higher \( s \) values promote retrieval diversity by prioritizing documents that may be less similar in their embeddings but more relevant from a broader perspective.
\begin{table}[t]
\centering
\begin{tabular}{ccc}
\toprule
\textbf{Parameter ($s$)} & $\tau$ & $\rho$ \\
\midrule
0.0 & 1.00 & 1.00 \\
0.2 & 0.797 & 0.828 \\
0.4 & 0.688 & 0.742 \\
0.6 & 0.485 & 0.528 \\
0.8 & 0.265 & 0.316 \\
1.0 & 0.074 & 0.078 \\
\bottomrule
\end{tabular}
\caption{Sensitivity Analysis of the VSR. Higher $s$ values indicate a greater degree of diversity introduced in the ranking by the retrieval process.}
\label{tab:sensitivity_analysis}
\end{table}
\paragraph{Vendi retrieval improves document retrieval diversity.} To assess the impact of the Vendi retrieval process on retrieval diversity, we compared the diversity of outputs from Vendi-RAG against Adaptive-RAG and Adaptive Retrieval. We measured diversity using two different metrics, the VS and the max pairwise distance (MPD). Table~\ref{tab:vs_mpd_comparison} summarizes the results. Vendi-RAG achieves higher diversity compared to Adaptive Retrieval and Adaptive-RAG on all dataset, demonstrating its ability to retrieve documents that capture multiple perspectives relevant to the query. This is a crucial advancement, as increased diversity in retrieval directly correlates with improved robustness in multi-hop reasoning (see Table \ref{tab:main:gpt}). Adaptive-RAG, which incorporates iterative refinement but lacks explicit diversity control, shows moderate retrieval diversity improvement over Adaptive Retrieval. 
\begin{table*}[t]
\small
\centering
\resizebox{\textwidth}{!}{
\renewcommand{\arraystretch}{1.0}
\begin{tabular}{lccccccccccc}
\toprule
 & \multicolumn{3}{c}{ MuSiQue} & \multicolumn{3}{c}{ HotpotQA} & \multicolumn{3}{c}{ 2WikiMultiHopQA} \\
\cmidrule(l{2pt}r{2pt}){2-4} \cmidrule(l{2pt}r{2pt}){5-7} \cmidrule(l{2pt}r{2pt}){8-10}
 {Methods} & EM & F1 & Acc & EM & F1 & Acc & EM & F1 & Acc \\
\midrule
 {No Retrieval} & 20.4 & 31.3 & 24.4 & 37.4 & 51.0 & 43.2 & 37.0 & 48.5 & 43.4 \\
{Single-step Approach} & 16.4 & 26.7 & 23.6 & 39.6 & 50.4 & 45.6 & 46.8 & 57.7 & 52.6 \\
{Adaptive Retrieval} & 18.8 & 30.3 & 24.8 & 38.6 & 50.7 & 43.2 & 44.2 & 55.1 & 50.6 \\
{LightRAG} & 17.4 & 27.5 & 23.8 & 37.4 & 47.3 & 42.3 & 42.2 & 52.2 & 47.5 \\
 {Adaptive-RAG} & 21.8 & 32.6 & 29.6 & 40.4 & 52.5 & 47.0 & 46.6 & \textbf{60.1} & 56.8 \\
 {Self-RAG} &  1.2 & 8.2 & 11.8 & 5.6 & 17.8 & 30.6 & 3.0 & 19.1 & 39.0 \\
 Adaptive Retrieval with MMR & 22.6 & 31.0 & 28.8 & 41.0 & 55.0 & 56.0 & 45.8 & 57.0 & 59.0 \\
 Graph-RAG & 22.9 & 32.5 & 30.1 & 42.4 & 57.2 & 57.8 & 47.5 & 59.1 & 60.9 \\
 FiD-Reranker  & 23.2 & \textbf{33.1} & 29.8 & 41.8 & 56.7 & 56.9 & 46.6 & 59.3 & 60.1 \\
 \hline
 Vendi-RAG(fixed-$s_{1:N}=0.0$) & 22.4 & 30.4 & 28.2 & 40.6 & 54.1 & 55.4 & 45.4 & 56.8 & 58.8 \\
Vendi-RAG$^*$(fixed-$s_{1:N}=0.3$) & 22.4 & 31.1 & 28.6 & 40.7 & 55.0 & 55.7 & 45.3 & 56.8 & 59.1 \\
Vendi-RAG(fixed-$s_{1:N}=0.8$) & {22.6} & 30.4 & {29.1} & {41.2} & {55.7} & {57.2} & {46.4} & 57.4 & {60.2} \\

Vendi-RAG(fixed-$s_{1:N}=0.8$) & {23.0} & 31.2 & {30.2} & {42.0} & {56.9} & {58.0} & {47.0} & 58.7 & {61.0} \\

Vendi-RAG(fixed-$s_{1:N}=1.0$) & 22.9 & 31.0 & 29.4 & 41.6 & 55.8 & 57.0 & 46.4 & 57.8 & 60.2 \\

\textbf{Vendi-RAG}($s_1=0.8$) & \textbf{24.4} & 32.5 & \textbf{30.4} & \textbf{42.2} & \textbf{57.0} & \textbf{58.4} & \textbf{47.2} & 58.9 & \textbf{61.4} \\

\bottomrule
\end{tabular}
}
\caption{Performance on multi-hop QA datasets using GPT-3.5 Turbo is evaluated across three metrics: exact match (EM), F1 score (F1), and traditional accuracy (Acc). Vendi-RAG with $s_1=0.8$ outperforms all baselines across the three datasets in terms of EM and Acc, while achieving comparable F1 scores to Adaptive-RAG. Here, Vendi-RAG$^*$ refers to the variant of Vendi-RAG that excludes the LLM-Judge component.}
\label{tab:main:gpt}
\end{table*}
\paragraph{Accuracy on multi-hop QA tasks.} We further evaluated the performance of the Vendi-RAG pipeline to assess its ability to reason across multiple documents. The results in Table~\ref{tab:main:gpt} indicate that Vendi-RAG consistently outperforms other methods in response accuracy across all datasets, showcasing the efficacy of balancing retrieval diversity with quality. Additionally, Vendi-RAG achieves competitive performance in Exact Match (EM) and F1 score. These findings highlight Vendi-RAG's capability to enhance answer correctness for complex reasoning tasks through improved document retrieval. Additionally, we conducted a comprehensive ablation study comparing various retrieval strategies, including Vendi-RAG and MMR, as well as a dynamic adjustment of $s_t$ against fixed $s_t$ in the table. The results demonstrate that Vendi-RAG with dynamically adjusted $s_t$ (with an initial setting of $s_1=0.8$) with the LLM-Judge consistently outperforms all baselines, including MMR, across the datasets in terms of EM, F1, and Acc.

\begin{figure*}[t]
\centering
\includegraphics[width=1\linewidth]{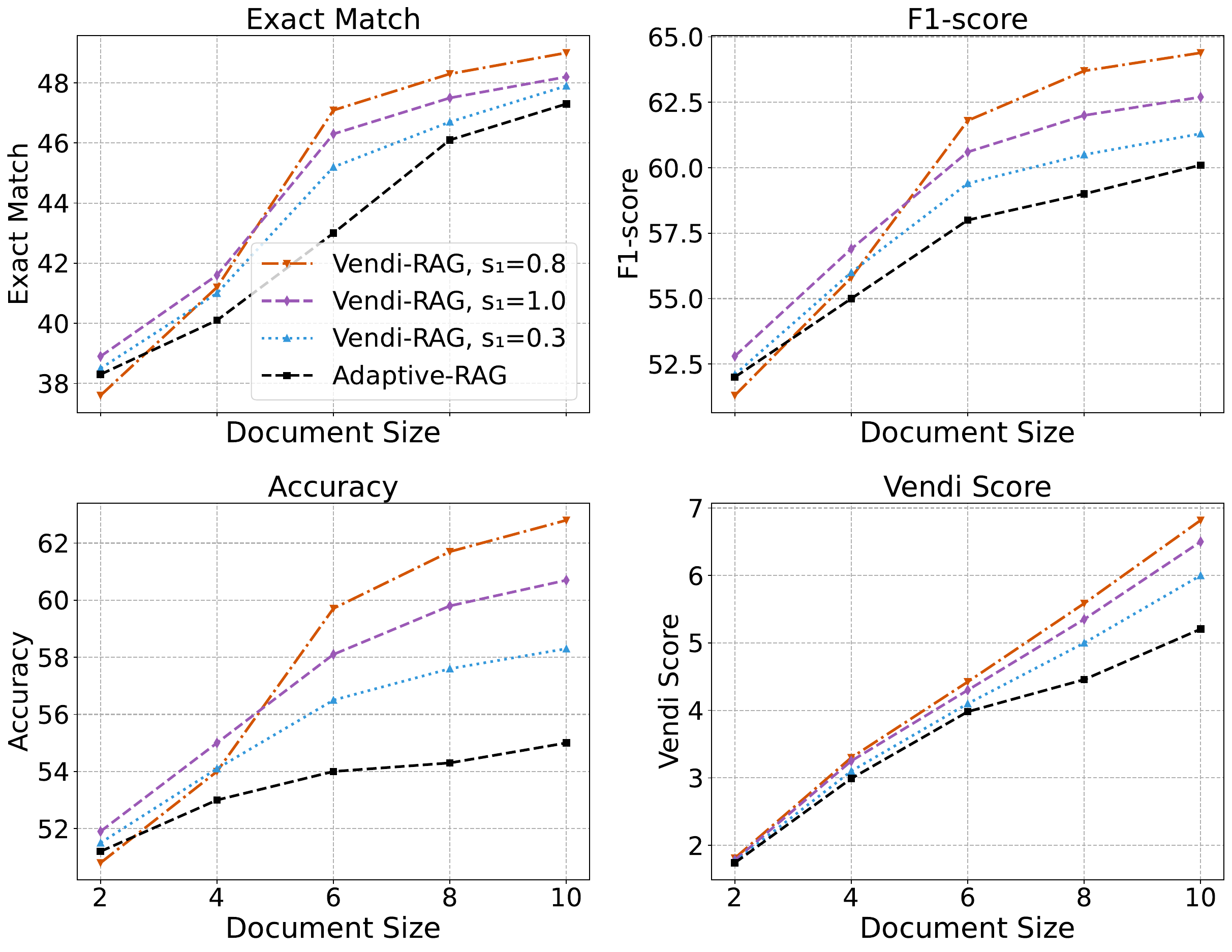}
\caption{Performance comparison of Vendi-RAG and Adaptive-RAG across different document sizes in terms of Exact Match, F1-score, Accuracy, and Vendi Score on HotPotQA. The backbone LLM used is \texttt{GPT-4o-mini}. Vendi-RAG consistently outperforms Adaptive-RAG across all metrics. In particular, performance improves as the number of retrieved documents increases. Different variants of Vendi-RAG are plotted based on the fixed initialization value \( s_1 \) for the diversity-relevance parameter \( s_t \), with \( s_1 = 0.8 \) achieving the best overall results.}
\label{fig:rag-doc_size}
\end{figure*}
\paragraph{Ablation Study on Retrieval Strategy and $s_t$ Scheduling.}
To better understand the effectiveness of our retrieval strategy, we conducted a comprehensive ablation study, examining various configurations of the Vendi-RAG pipeline. First, we evaluated fixed values of $s_t$ across all steps, testing settings such as $s_t=\{0.0, 0.3, 0.8, 1.0\}$. Among the fixed schedules, setting $s_t=0.8$ consistently achieved the best overall performance in terms of Exact Match (EM), F1 score, and Accuracy across all datasets, highlighting the importance of dynamically balancing retrieval diversity with quality. We further compared Vendi-RAG against traditional retrieval baselines, including Adaptive Retrieval with MMR, and observed that our method outperformed the MMR retrieval method across all metrics. Additionally, we tested a variant without dynamic scheduling (fixed $s_t$) and a variant without the LLM-Judge module (Vendi-RAG$^*$). The results show that dynamically adjusting $s_t$ during retrieval using the LLM-Judge significantly boosts performance compared to fixed schedules and simpler retrieval strategies. These findings emphasize the critical role of adaptive retrieval and document assessment in enabling Vendi-RAG to effectively handle complex multi-hop reasoning tasks.
\paragraph{Impact of the number of retrieved documents on performance.} To further examine the impact of document size on retrieval effectiveness, we analyze the performance of Vendi-RAG and Adaptive-RAG across varying document sizes and initial settings of $s_1=\{0.3,0.8,1.0\}$. Figure~\ref{fig:rag-doc_size} illustrates the relationship between document size and performance on the HotPotQA dataset. Vendi-RAG consistently outperforms Adaptive-RAG in accuracy for document sizes greater than two with any $s_1$ setting. As document size increases, accuracy improves for both methods, but the gain is notably higher for Vendi-RAG. Similar to accuracy, EM and F1 scores exhibit an increasing trend as document size grows. Vendi-RAG shows a more pronounced improvement, underscoring its capacity to retrieve more informative and relevant documents, thereby enhancing answer quality. The VS also increases with document size. This is evidence that Vendi-RAG alleviates redundancy since the VS is known to be invariant under duplication~\citep{friedman2023vendi}. An increasing VS indicates less redundancy in the retrieved documents. By leveraging the VS in its retrieval process, Vendi-RAG avoids the redundancy issue that often plagues RAG pipelines. These results indicate that increasing document size enhances both retrieval diversity and answer correctness. Vendi-RAG is achieving superior gains in all metrics. However, we are computationally bottlenecked primarily by the LLM's context window limitation and processing time. As the number of retrieved documents increases, we must either truncate documents to fit within the model's maximum context length or process documents in multiple batches, both of which have significant computational overhead.
\paragraph{Performance for different LLM Backbones and retrieval strategies.}
\begin{figure}[tb]
\centering
\includegraphics[width=.95\linewidth]{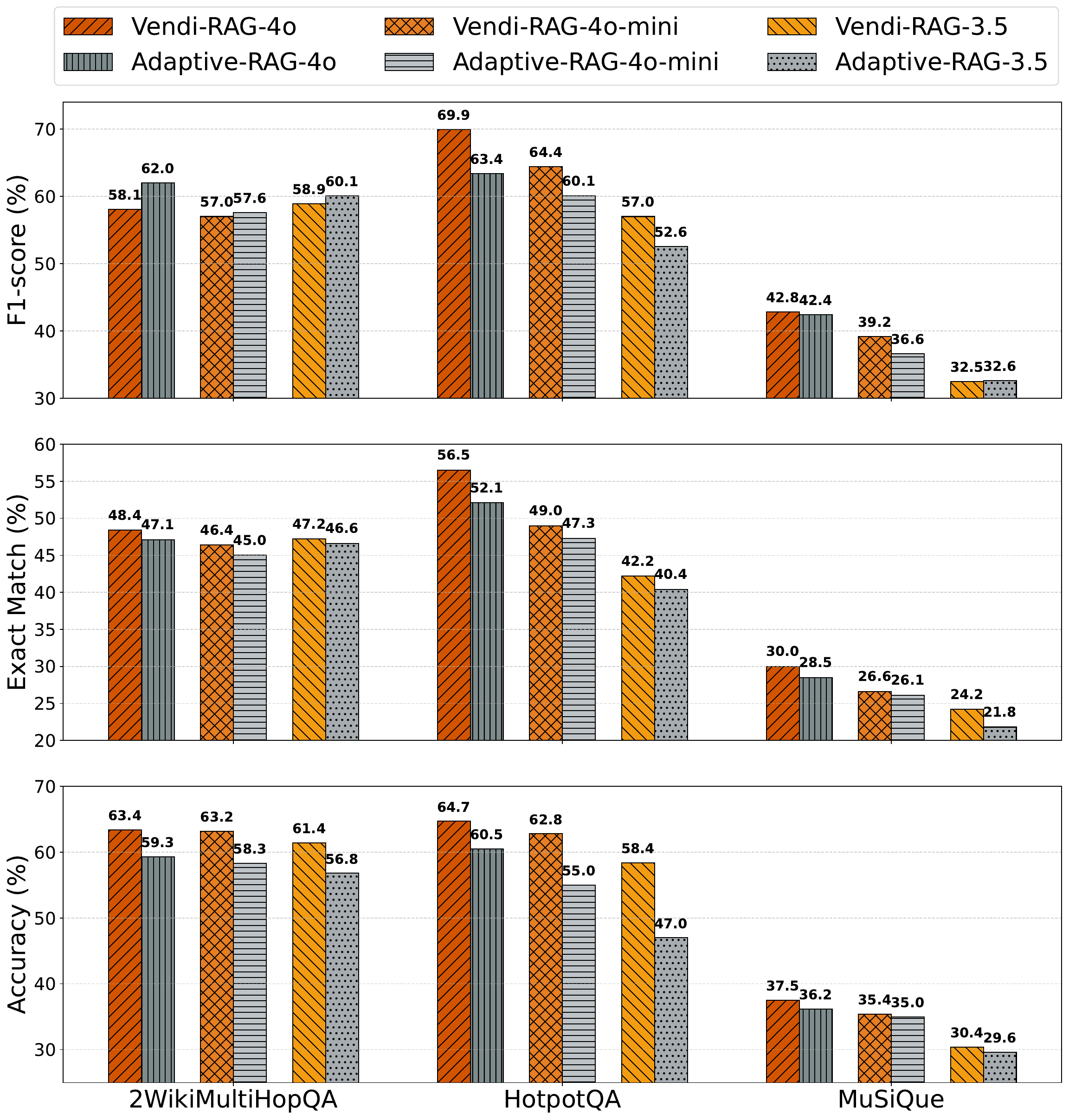}
\caption{Performance comparison of Vendi-RAG and Adaptive-RAG variants across the three datasets using three evaluation metrics: F1-score, Exact Match, and Accuracy. Results show that Vendi-RAG-4o consistently outperforms other variants across all datasets and metrics, with a particularly strong performance on HotpotQA.}
\label{fig:rag-llm-comparison}
\end{figure}
To evaluate the impact of different LLM backbones and retrieval strategies on the performance of the Vendi-RAG (with $s_1=0.8$) framework, we conducted experiments using three LLMs: GPT-4o, GPT-4o-mini, and GPT-3.5, across all the multi-hop QA datasets described above. The results, shown in Figure~\ref{fig:rag-llm-comparison}, highlight the effectiveness of Vendi-RAG compared to Adaptive-RAG, the best baseline, across all datasets and LLM-backbones, except for F1-score on the 2WikiMultiHopQA dataset. In general, larger LLM backbones, such as GPT-4o, achieve superior performance across all datasets, particularly for tasks requiring complex reasoning and synthesis across multiple documents. However, even with smaller models like GPT-4o-mini, the Vendi-RAG model maintains competitive performance, demonstrating its effectiveness. 
\paragraph{VSR vs MMR.}
VS offers key advantages over traditional diversity metrics like MMR. While MMR relies on pairwise comparisons to balance relevance and novelty, it lacks a global view of semantic diversity across the retrieved set. In contrast, VS is a principled, global metric based on the eigenvalues of the normalized kernel matrix, directly measuring the effective number of distinct documents. It reaches its maximum when documents are entirely unique and its minimum when they are identical, providing an intuitive, mathematically grounded measure of diversity. This global perspective makes VS particularly effective for multi-hop QA, where broad semantic coverage is critical. Moreover, VS integrates naturally with Vendi-RAG's dynamic retrieval adjustment, enabling fine-grained control over the diversity-relevance trade-off via a single parameter, and addressing the challenge of balancing coverage and precision in complex reasoning tasks.

\section{Conclusion}
\label{sec:conclusion}

While RAG has proven effective in enhancing LLM performance for domain-specific QA tasks, traditional models often struggle with redundancy, particularly in multi-hop reasoning tasks. To address this shortcoming, we introduce Vendi-RAG, a novel framework that jointly optimizes retrieval diversity and answer quality through an iterative refinement process. Vendi-RAG leverages the Vendi Score and an LLM judge to promote semantic diversity while maintaining relevance during retrieval. Our experiments on HotpotQA, MuSiQue, and 2WikiMultiHopQA demonstrate Vendi-RAG's effectiveness. Specifically, Vendi-RAG outperforms the best baseline by $+4.2\%$ on HotpotQA, $+4.1\%$ on 2WikiMultiHopQA, and $+1.3\%$ on MuSiQue. These gains become even more pronounced as the number of retrieved documents increases, highlighting the importance of retrieval diversity in complex reasoning tasks. Furthermore, we evaluated Vendi-RAG across multiple LLM backbones, including GPT-3.5, GPT-4, and GPT-4o-mini, and observed consistent performance improvements, demonstrating that the framework is model-agnostic. These findings establish Vendi-RAG as an effective and adaptable solution for multi-hop QA. 

\section{Limitations}
Vendi-RAG introduces computational overhead due to LLM-based quality scoring, which may limit scalability in real-time applications. Additionally, like all RAG approaches, its performance depends on the quality and completeness of external knowledge sources, making it susceptible to biases or gaps in the retrieved information.

\section{Ethics Statement}
The deployment of LLMs, including their use in Vendi-RAG, necessitates careful ethical consideration. Since the model relies on external knowledge sources, concerns arise regarding the credibility and accuracy of retrieved content. Ensuring the reliability and factual integrity of information is crucial to mitigating risks related to bias and misinformation.

\subsection*{Acknowledgements}
Adji Bousso Dieng acknowledges support from the National Science Foundation, Office of Advanced Cyberinfrastructure (OAC): \#2118201. She also acknowledges Schmidt Sciences for the AI2050 Early Career Fellowship.

\bibliographystyle{apa}
\bibliography{arxiv}

\appendix

\section{Datasets}
\label{app:datasets}
MuSiQue evaluates a model’s ability to synthesize facts spread across across multiple document sources. It includes questions spanning diverse domains such as history, science, and culture, requiring logical reasoning and synthesis of interdependent information. Given its emphasis on multi-step comprehension, this dataset challenges models to accurately identify and integrate relevant information to generate correct answers to queries. This is the most challenging dataset among the three.

HotpotQA assesses reasoning and fact verification across various domains, including geography, entertainment, and history. Its questions necessitate reasoning over two or more interconnected documents linked via hyperlinks. Additionally, the dataset includes “comparison” questions that require juxtaposing information from multiple sources, making it a challenging benchmark for evaluating both retrieval quality and reasoning ability.

2WikiMultiHopQA leverages Wikipedia’s complex structure to pose complex reasoning challenges. Questions are derived from real-world knowledge graphs and require navigating reasoning paths across multiple documents. Topics span science, politics, and sports, emphasizing logical relationships such as cause-effect dependencies, making it an essential tool for evaluating structured knowledge reasoning.
\section{Evaluation Metrics}
\label{app:eval_metrics}
To compare model performance across different datasets, we employ the following key evaluation metrics:

\begin{itemize}
    \item \textbf{Exact Match (EM)}: This metric calculates the percentage of predictions that exactly match the ground truth answers. It is defined as:
    \begin{equation}
    \text{EM} = \frac{\text{Number of exact matches}}{\text{Total number of queries}} \times 100
    \end{equation}
    EM is a strict metric that grants credit only when the predicted answer matches the ground truth exactly in both content and format. It is particularly useful for assessing a model's precision in generating accurate responses.

    \item \textbf{F1 Score (F1)}: The F1 score captures the harmonic mean of precision and recall at the token level, providing a balanced measure of correctness. It is defined as:
    \begin{equation}
    \text{F1} = 2 \cdot \frac{\text{Precision} \cdot \text{Recall}}{\text{Precision} + \text{Recall}}
    \end{equation}
    where precision is the fraction of retrieved tokens that are relevant, and recall is the fraction of relevant tokens that are retrieved. The F1 score is particularly relevant for multi-hop QA tasks, where partial correctness (e.g., retrieving some but not all supporting evidence) is informative.

    \item \textbf{Accuracy (Acc)}: Accuracy measures the proportion of correct predictions over all evaluated queries. It is defined as:
    \begin{equation}
    \text{Acc} = \frac{\text{Number of correct predictions}}{\text{Total number of queries}} \times 100
    \end{equation}
    Unlike EM, which requires exact matches, accuracy provides a broader assessment by capturing overall correctness, including responses that convey the intended meaning.

    \item \textbf{Max Pairwise Distance (MPD)}: This metric evaluates the maximum Euclidean distance between pairs of retrieved data points, measuring diversity. It is defined as:
    \begin{equation}
    \text{MPD} = \max_{i, j} \|x_i - x_j\|_2, \quad i \neq j
    \end{equation}
    where \(x_i\) and \(x_j\) represent document embeddings in the feature space. Higher values indicate greater diversity among retrieved documents.
\end{itemize}
Each of these metrics offers a unique perspective on model performance. EM is a stringent measure of precision, F1 balances precision and recall, and accuracy provides an overall correctness measure. Meanwhile, MPD and diversity-based metrics assess the variety and independence of retrieved documents, critical for multi-hop QA tasks requiring integration of diverse information.

\section{Implementation Details} 
The Vendi-RAG framework employs dense vector representations derived from transformer-based embeddings to compute the similarity score $ \text{SS}(q, \mathcal{D})$ between a query $q$ and a set of documents $\mathcal{D}$. This high-dimensional semantic comparison allows the model to effectively capture contextual relationships and retrieve relevant documents across diverse domains.
To evaluate answer quality, we use a consistent LLM backbone with a specifically designed prompt that positions the LLM as an expert judge. This judge-based evaluation framework assesses the generated answers according to this prompt:

\begin{tcolorbox}[colback=gray!10, colframe=black, title=LLM as a Judge Prompt]
\label{pr:judge}
\textbf{You are an expert LLM-based judge tasked with evaluating the quality of answers in a Retrieval-Augmented Generation (RAG) system. Your evaluation will consider the following aspects:\\}

1. \textbf{Coherence}: Assess whether the provided answer is logically consistent and flows smoothly, without conflicting statements or gaps in reasoning.  

2. \textbf{Relevance}: Evaluate how well the answer addresses the query based on the information from the retrieved documents.  

3. \textbf{Query Alignment}: Determine how closely the answer aligns with the specific query asked, ensuring that the response is focused and appropriate.  \\

\textbf{Your evaluation will be quantified based on the following scoring system:\\}  
- Coherence Score (C): [1 - 10], where 10 is perfectly coherent.  

- Relevance Score (R): [1 - 10], where 10 is highly relevant to the query.  

- Query Alignment Score (Q): [1 - 10], where 10 is perfectly aligned.  

Provide a quality score \( Q_t \) as the average of these individual scores:  
\[
Q_t = \text{mean}(C, R, Q)
\]
Query: \{query\}
\end{tcolorbox}

This judge assesses coherence, relevance, and alignment with the query to produce a quality score \( Q_t \). The quality threshold (\( \tau \)) is set to 0.85 for all experiments, ensuring a consistent standard of answer evaluation. While we initially set \( s_1 = 0.8 \) across experiments to prioritize diversity, our ablation study explores this hyperparameter through testing different fixed values and employing dynamic adjustment. This analysis highlights the importance of dynamic adjustment for improved performance across diverse datasets.

\subsection{Dataset Processing and Chunking}
\label{app:vector_db}
Preparing datasets for question-answering requires transforming data into a searchable vector database to enable efficient retrieval. This workflow includes document chunking and semantic embedding to optimize performance.

The dataset, provided in JSON format with context paragraphs and metadata, is processed by splitting each document into smaller chunks. Each chunk has a maximum size of 512 tokens, with a 50-token overlap to preserve context across chunk boundaries and facilitate multi-hop reasoning in long documents.

\subsection{Embedding Model and Vector Database}
We use the \textbf{SentenceTransformer} model, specifically \texttt{all-mpnet-base-v2}, to generate dense vector representations for documents and queries. These embeddings are stored locally to avoid redundant downloads and improve reusability. The \textbf{Chroma} vector database efficiently stores and retrieves these vectorized documents along with metadata, such as document titles and chunk IDs.

\subsection{Batch Processing and Database Population}
To efficiently populate the vector database, document chunks are processed in batches of up to 10,000. This approach optimizes memory usage while ensuring completeness in the ingestion process. The total number of processed chunks is logged for verification.

\subsection{Query Answering Workflow}
For queries such as \textit{"Who is the father-in-law of Queen Hyojeong?"}, relevant chunks are retrieved using \textbf{Chroma}'s similarity-based search mechanism. The system ranks the top 10 chunks based on their semantic similarity to the query, leveraging embeddings generated by \texttt{all-mpnet-base-v2} to ensure precise and relevant results.

\subsection{Key Configuration Details}
The system is configured with the following parameters:
\begin{itemize}
    \item \textbf{Embedding Model:} \texttt{all-mpnet-base-v2}, optimized for semantic similarity.
    \item \textbf{Vector Database:} Chroma, persisted to disk for efficient reuse.
    \item \textbf{Chunk Size:} 512 tokens per chunk, with a 50-token overlap.
    \item \textbf{Batch Size:} Up to 10,000 chunks per batch.
\end{itemize}

\end{document}